\documentclass{article}

\usepackage{amsmath}
\usepackage{amssymb}
\usepackage{amsthm}
\usepackage{array,makecell} 
\usepackage{booktabs}
\usepackage{caption}
\usepackage{float}
\usepackage{gensymb}
\usepackage{graphicx}
\usepackage{lineno}
\usepackage{mathtools}
\usepackage{microtype}
\usepackage{multirow,bigdelim}
\usepackage{soul}
\usepackage{subfigure}
\usepackage{xspace}
\usepackage[textsize=tiny]{todonotes}
\usepackage{hyperref}
\usepackage[capitalize,noabbrev]{cleveref}

\usepackage[accepted]{icml2024}

\theoremstyle{plain}

\theoremstyle{definition}

\theoremstyle{remark}

\usepackage[textsize=tiny]{todonotes}

\icmltitlerunning{DATAP-SfM: Dynamic-Aware Tracking Any Point for Robust Structure from Motion in the Wild}

\begin{document}
{
\twocolumn[{
\icmltitle{DATAP-SfM: Dynamic-Aware Tracking Any Point \\ for Robust Structure from Motion in the Wild}





\begin{icmlauthorlist}
\icmlauthor{Weicai Ye}{zju}
\icmlauthor{Xinyu Chen}{zju}
\icmlauthor{Ruohao Zhan}{zju}
\icmlauthor{Di Huang}{shailab}
\icmlauthor{Xiaoshui Huang}{shailab}
\icmlauthor{Haoyi Zhu}{shailab}
\icmlauthor{Hujun Bao}{zju}
\icmlauthor{Wanli Ouyang}{shailab}
\icmlauthor{Tong He}{shailab}
\icmlauthor{Guofeng Zhang}{zju}
\end{icmlauthorlist}

\icmlaffiliation{zju}{State Key Lab of CAD\&CG, Zhejiang University}
\icmlaffiliation{shailab}{Shanghai AI Laboratory}

\icmlcorrespondingauthor{Weicai Ye}{maikeyeweicai@gmail.com}
\icmlcorrespondingauthor{Guofeng Zhang}{zhangguofeng@zju.edu.cn}




\icmlkeywords{Machine Learning, ICML}
\vskip 0.3in

\begin{center}
  \includegraphics[width=\textwidth]{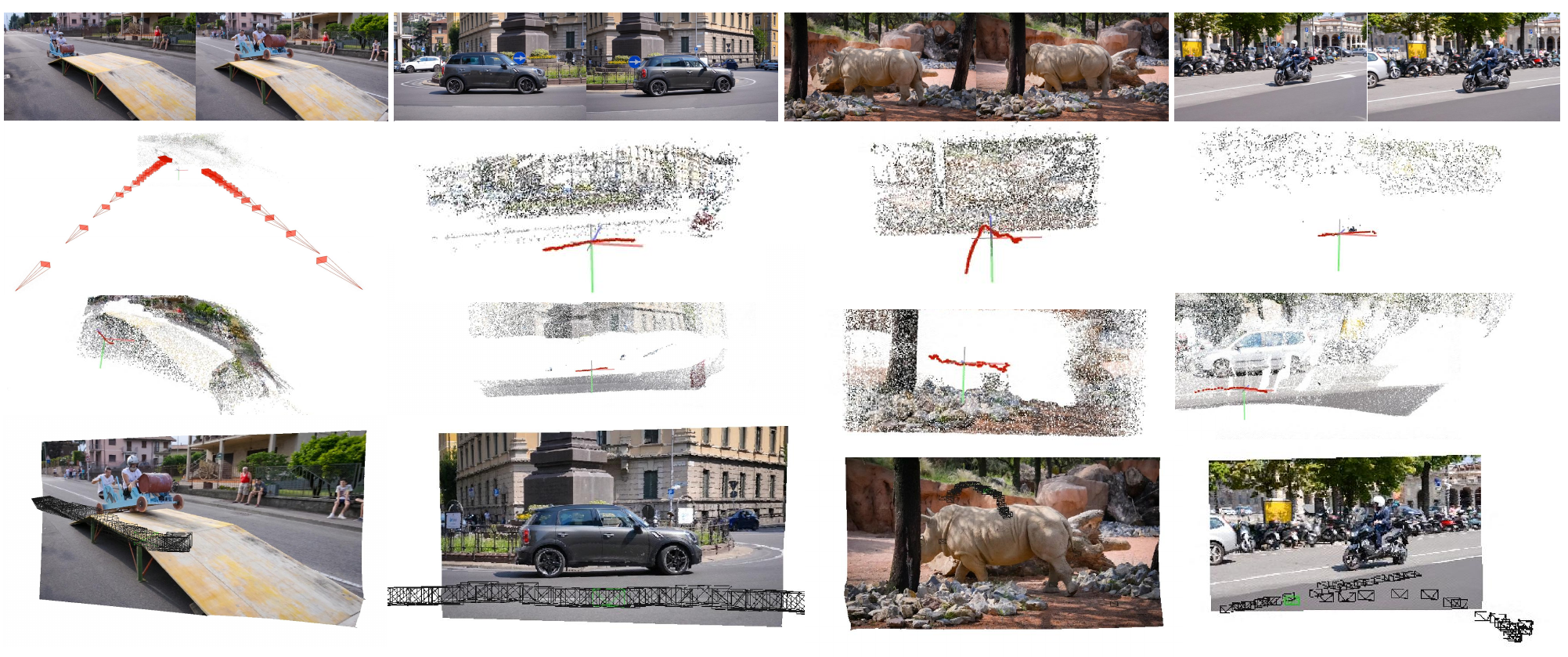}
  \captionof{figure}{\textbf{Given casual videos, our method can obtain smooth camera trajectories and entire point clouds of dynamic scenes.} From top to bottom: video samples, results from COLMAP, ParticleSfM, and ours.}
  \label{fig:teaser}  
  
\end{center}
}
]



\printAffiliationsAndNotice{}  

\begin{abstract}
This paper proposes a concise, elegant, and robust pipeline to estimate smooth camera trajectories and obtain dense point clouds for casual videos in the wild. Traditional frameworks, such as ParticleSfM~\cite{zhao2022particlesfm}, 
address this problem by sequentially computing the optical flow between adjacent frames to obtain point trajectories. They then remove dynamic trajectories through motion segmentation and perform global bundle adjustment.
However, the process of estimating optical flow between two adjacent frames and chaining the matches can introduce cumulative errors. Additionally, motion segmentation combined with single-view depth estimation often faces challenges related to scale ambiguity.
To tackle these challenges, we propose a dynamic-aware tracking any point (DATAP) method that leverages consistent video depth and point tracking. Specifically, 
our DATAP addresses these issues by estimating 
dense point tracking across the video sequence and predicting the visibility and dynamics of each point. By incorporating the consistent video depth prior, the performance of motion segmentation is enhanced. With the integration of DATAP, it becomes possible to estimate and optimize all camera poses simultaneously by performing global bundle adjustments for point tracking classified as static and visible, rather than relying on incremental camera registration. Extensive experiments on dynamic sequences, e.g., Sintel and TUM RGBD dynamic sequences, and on the wild video, e.g., DAVIS, demonstrate that the proposed method achieves state-of-the-art performance in terms of camera pose estimation even in complex dynamic challenge scenes.

\end{abstract}

\section{Introduction}
Estimating the pose of moving cameras from monocular videos plays a fundamental role in computer vision and robotics, finding applications in various fields such as autonomous driving and augmented reality. 
In everyday casual videos, 
the camera is typically moving, while complex foreground movements, including people, vehicles, and other moving objects, dominate the majority of the video frames. This introduces significant challenges in achieving robust camera pose estimation in such scenarios.

Traditional indirect SLAM~\cite{forster2014svo, orbslam3, ptam} or SfM~\cite{schoenberger2016sfm} methods 
extract and match high-quality feature points and utilize nonlinear optimization techniques to estimate camera poses and reconstruct 3D point clouds by minimizing geometric reprojection errors. In contrast, direct SLAM~\cite{dso, lsd, dtam} or SfM methods perform camera tracking by optimizing photometric errors, assuming consistent video appearance. 
While these methods have shown promising results, they often struggle with robust localization in scenes that contain a significant number of dynamic objects. This limitation becomes particularly evident in real-world scenarios where dynamic objects are common.

To solve this problem, some visual odometry~\cite{svo, kim2019simvodis, Ye2023PVO} or SLAM~\cite{bescos2018dynaslam, xiao2019dynamic} methods use semantic or geometry~\cite{Wei2013robust} priors to mitigate the interference caused by specific types of potential dynamic objects, such as humans or vehicles. However, in practical natural scenes, some seemingly static objects may also exhibit motion, such as water cups being picked up or moved, or willow branches swaying in the wind. This renders the aforementioned methods relying on semantic constraints ineffective. On the other hand, some end-to-end visual odometry~\cite{d3vo, li2021unsupervised, tartanvo} or SLAM methods~\cite{teed2021droid, Ye2022DeFlowSLAM} implicitly model the complex motion of scene objects and estimate camera pose by focusing on static areas through training data. However, these methods face challenges when generalizing to wild videos.

Recently, ParticleSfM~\cite{zhao2022particlesfm} 
introduced a method that involves constructing point trajectories, applying trajectory motion segmentation to address dynamic trajectory interference, and performing global bundle adjustment for improved pose estimation.
This method has shown promise in terms of generalizing well and accurately estimating poses, yet it comes with notable limitations: 
(1) Point trajectory construction based on pairwise optical flow matching will undoubtedly bring \emph{long-term cumulative errors}. 
(2) Motion segmentation using monocular pose estimation suffers from \emph{scale ambiguity}.

Building on the impressive performance of recent 2D point tracking methods such as TAPIR~\cite{doersch2023tapir}, CoTracker~\cite{karaev2023cotracker} and Omniotion~\cite{wang2023omnimotion}, we introduce a novel approach called Dynamic-Aware Tracking Any Point (DATAP) to address these drawbacks by leveraging consistent video depth estimation~\cite{nvds} and long-term point tracking~\cite{karaev2023cotracker}.
Specifically, DATAP is a transformer network that operates in a sliding window fashion. It estimates the point tracking and visibility of the sampled points across the videos. 
The transformer network incorporates self-attention and cross-attention mechanisms, treating each trajectory within the sliding window as a whole. This allows for the exploitation of correlation among trajectory features and facilitates the exchange of information within and between trajectories.
Within the sliding window, the trajectory of each query point is initially set to 0. The network will progressively refine these initial estimates through the iteration of the transformer. 
Subsequent overlapping windows initialize the trajectory and visibility based on the refinement predictions from the previous window, with updates made to the trajectory and visibility of the new frame.

To estimate the dynamic motion label of a trajectory, we have incorporated a multi-layer perceptual layer into the tracked features. This additional layer predicts the probability of dynamic motion, similar to how visibility is predicted. By doing so, we aim to address the ambiguity that can arise when using 2D point tracking alone for dynamic prediction.
Inspired by ParticleSfM~\cite{zhao2022particlesfm}, we leverage depth information and design a transformer module to eliminate the ambiguity in 2D point motion segmentation. Considering the scale ambiguity posed by single-view depth estimation of monocular videos, such as Midas~\cite{Ranftl2022}, we propose to leverage consistent video depth estimation for depth initialization.

Incorporating with DATAP, we construct a concise, elegant, and robust pipeline for structure from motion in the wild
Experiments on MPI Sintel Dataset~\cite{ButlerSintel2012} and TUM RGBD dynamic sequences~\cite{tumrgbd} show that our structure from motion with dynamic-aware point tracking method can effectively improve the accuracy of camera localization in dynamic scenes. 
We also verified our method on casual videos in the wild such as DAVIS~\cite{davis}, demonstrating its robustness of localization in complex challenging scenarios. Overall, our contributions are summarized as follows. 

\begin{itemize}
    \item We propose a novel network of dynamic-aware tracking any point (DATAP) to simultaneously estimate point tracking, visibility, and dynamics from arbitrary videos in a sliding window manner, and exploit consistent video depth priors to further improve performance. 
    \item Incorporating DATAP and global bundle adjustment, we present a concise, elegant, and robust pipeline for smooth camera trajectories and dense point clouds from casual monocular videos.
    \item Extensive experiments demonstrate that the proposed method outperforms SOTA methods in complex dynamic challenge scenarios.
\end{itemize}

\section{Related Work}

\textbf{Correspondence Learning.} Estimating correspondences from two-view or multi-view videos is fundamental to many vision tasks. Correspondence relationships can be roughly divided into three representations: feature matching~\cite{superpoint, superglue}, optical flow estimation~\cite{raft, xu2022gmflow, ilg2017flownet} and point tracking~\cite{harley2022particle, karaev2023cotracker, doersch2023tapir, wang2023omnimotion}. Feature matching generally detects and matches high-quality feature points to cope with the challenges of large viewpoint changes and illumination variance. The optical flow estimation method finds the correspondence between adjacent frames based on the consistent video appearance. Obviously, methods based on feature matching and optical flow estimation will accumulate errors due to estimation deviations in video applications, causing drift. The point tracking method naturally acts on the video and jointly optimizes the point trajectory estimation of the entire sliding window, thereby better-reducing drift. Our dynamic-aware point-tracking method builds on advanced point tracking to jointly estimate trajectory, visibility, and dynamic properties in videos, enabling robust localization in complex and challenging dynamic scenes.

\begin{figure*}[htbp]
    \centering
    \includegraphics[width=\linewidth]{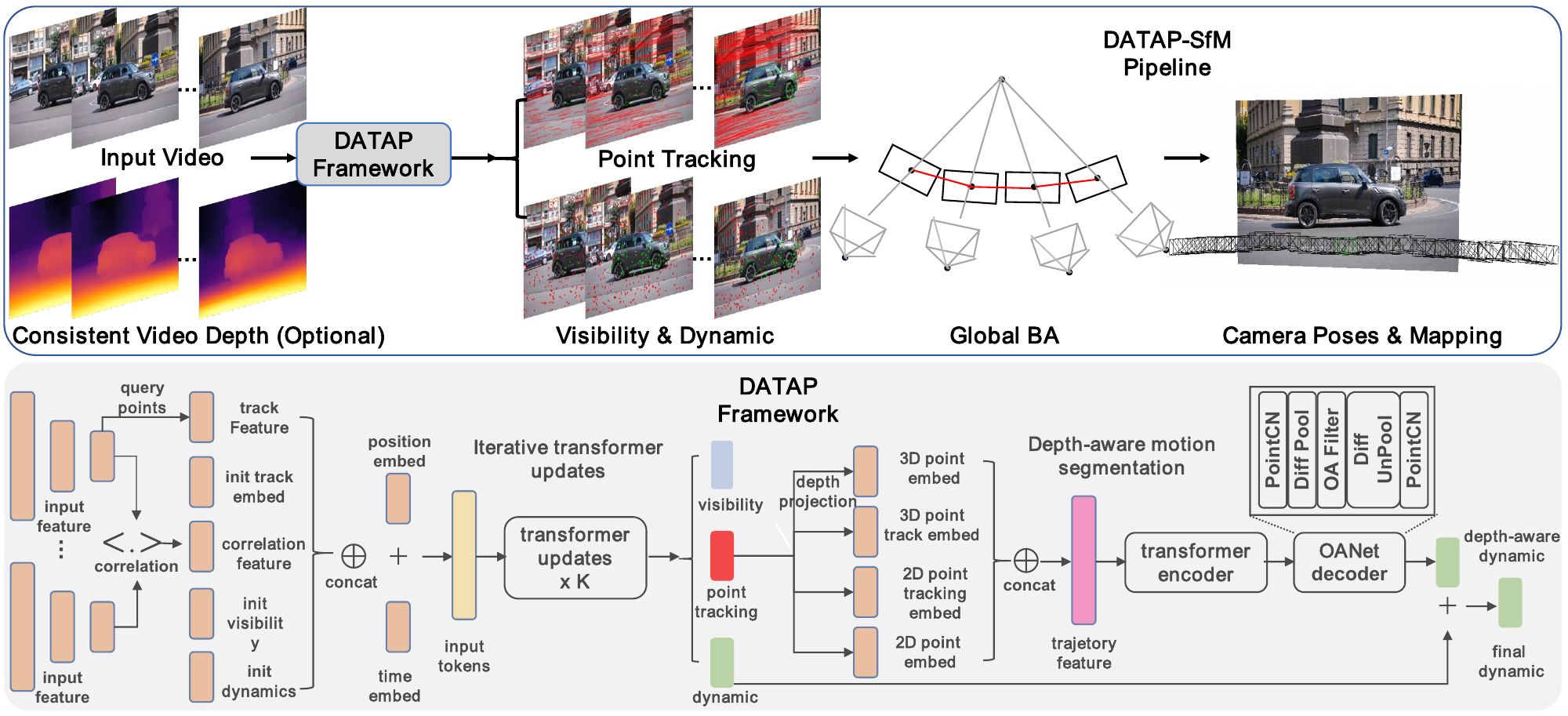}
    \caption{\textbf{DATAP-SfM pipeline.} Given monocular videos as input with consistent video depth (optional), DATAP can simultaneously estimate long-term point tracking with visible and dynamic characteristics. Incorporating with DATAP, we formalize a concise, elegant, and robust pipeline of structure from motion by performing global bundle adjustment for point tracking classified as static and visible.}
    \label{fig:overview}
\end{figure*}

\textbf{SLAM and Structure from Motion.} 
Localization and reconstruction play significant roles in computer vision and robotic applications and can be roughly divided into simultaneous localization and mapping (SLAM)~\cite{orbslam2, dso} (focusing on real-time positioning) and structure from motion(SfM)~\cite{colmap} (focusing on offline reconstruction). Traditional SLAM methods can be roughly divided into indirect methods~\cite{orbslam3} and direct methods~\cite{dso}. Among them, indirect methods generally perform feature extraction, feature matching, and outlier removal, and obtain accurate poses through nonlinear optimization of geometric reprojection errors. In contrast, the direct methods achieve this goal by optimizing photometric errors. Recently, many SLAM~\cite{teed2021droid} and visual odometry (VO)~\cite{deepvo} methods that combine traditional methods with deep learning have emerged, demonstrating compelling pose estimation results. To cope with the challenge of dynamic scenes, most methods~\cite{svo, Ye2023PVO} use semantic priors to remove the interference of potential dynamic objects of specific categories. 
SfM methods can be roughly divided into incremental methods~\cite{colmap} and global optimization methods~\cite{cui2015global}. Among them, incremental methods are generally faster, and global optimization methods are more accurate. 

Recently, ParticleSfM~\cite{zhao2022particlesfm} proposed to use series paired optical flows to obtain point trajectories, and then point trajectory motion segmentation and global BA methods can better cope with dynamic scenes. However, the method of chaining paired optical flows will undoubtedly bring cumulative errors in trajectory estimation, and the dynamic segmentation method based on single-view depth estimation will also be limited by the scale ambiguity of the video depth. Our method directly adopts advanced video point tracking methods and utilizes consistent video depth estimation, which can effectively solve the problems faced by ParticleSfM.

\textbf{Motion Segmentation.}
Motion segmentation aims to identify whether pixels in a video sequence are moving. Classic methods~\cite{shi1998motion, wang1994representing} use optical flow estimation to estimate motion labels and subsequent works~\cite{brox2006variational} jointly optimize optical flow estimation and motion segmentation to achieve better performance. Neural network-based methods simultaneously extract features and optical flow estimates or practical semantic constraints to group optical flows. Some methods~\cite{d3vo} perform joint optimization by simultaneously estimating camera motion, semantic information, and dynamic segmentation. 

Recently, some self-supervised methods~\cite{yang2021self} have emerged to alleviate the dependence on annotated data. 
However, optical flow estimation based on adjacent frames cannot capture long-term spatiotemporal features and is difficult to generalize to long videos. 
ParticleSfM proposed to use concatenating adjacent frames to obtain long-term point trajectories, and then use point trajectory motion segmentation to identify dynamic trajectories. 

It can generalize well to videos in the wild. However, concatenating adjacent frames to obtain long-term point trajectories will bring cumulative errors. Our method directly models point tracking and motion segmentation within a sliding window to eliminate the impact of cumulative errors.

\textbf{Single-View and Video Depth Estimation.} 
Depth estimation is an important intermediate representation for three-dimensional vision. Supervised deep learning methods~\cite{eigen2014depth, eigen2015predicting} require a large amount of dataset with ground truth labels. To solve these problems, some methods~\cite{megadepth} try to train depth estimation models on large-scale synthetic data or use multi-view stereo to obtain pseudo labels and achieve impressive results. However, single-view depth estimation methods can cause flickering problems in multi-view images or videos, so some methods~\cite{zhang09cvd} resort to using geometric and appearance-consistent priors between consecutive frames to deal with these challenges. They either use the geometric warping method to emphasize the consistency of the epipolar geometry of adjacent frames, or use joint optimization~\cite{rcvd} or finetune methods~\cite{cvd} to online adjust the results of single-view depth estimation. 

However, when faced with dynamic scenarios, dependencies such as geometric constraints will be broken, thus falling into local optima. Recently, NVDS~\cite{nvds} attempted to utilize large-scale stereo videos and learn the correlation between features to stabilize depth estimation across different frames and achieve consistent video depth estimation results. Our method takes advantage of consistent video depth to improve motion segmentation across videos and to achieve better pose estimation of monocular videos in the wild.

\begin{figure*}[htbp]
    \centering
    \includegraphics[width=\linewidth]{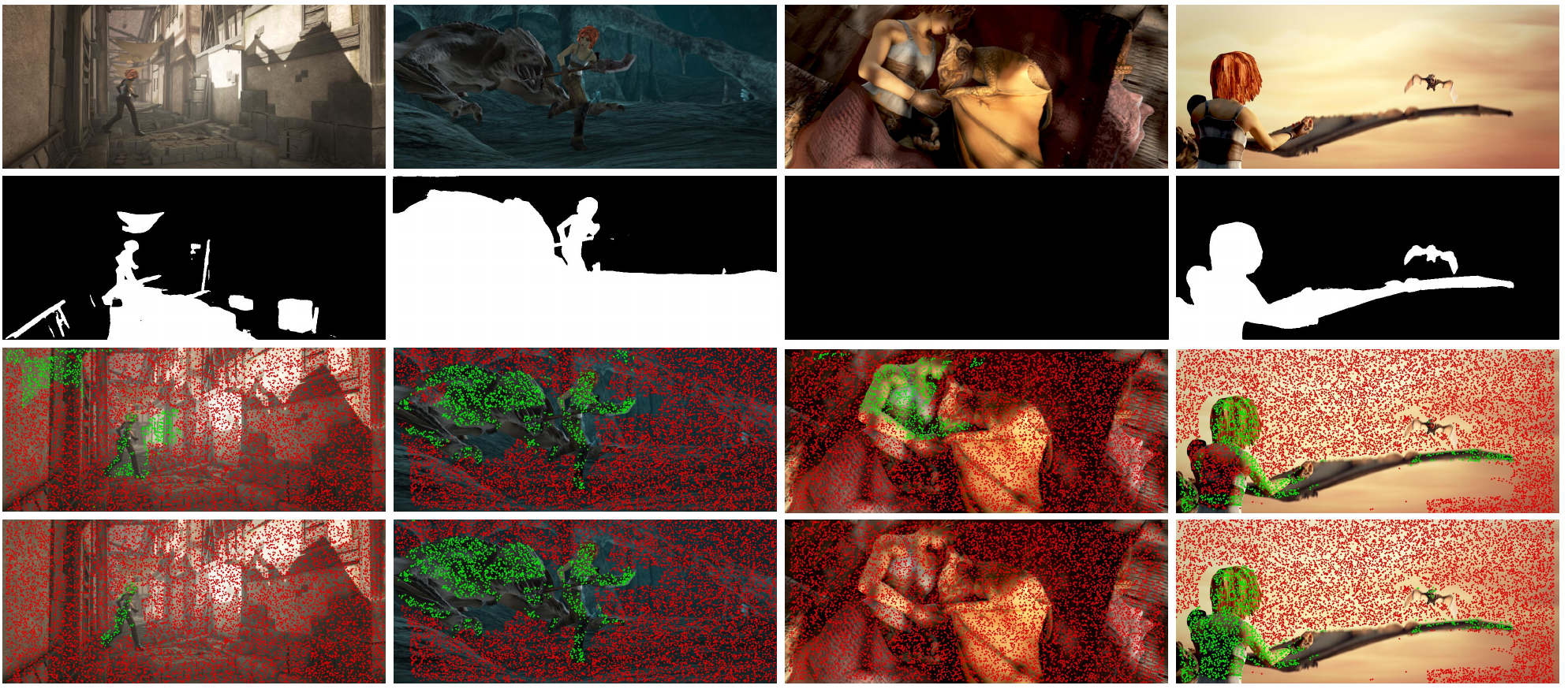}
    \caption{\textbf{Qualitative results of motion segmentation on MPI Sintel dataset. Our method outperforms existing SOTA methods. From top to bottom: image samples, motion segmentation results from Oneformer~\cite{jain2023oneformer}, ParticleSfM, and ours. Red: static, green: dynamic. The third column is the sleeping case, which should be static.}}
    \label{fig:sintel_motion}
\end{figure*}

\begin{table*}[t]
\centering
\caption{\textbf{Ablation study of motion segmentation on the MPI Sintel benchmark.} 
  }
\begin{tabular}{lcccc}
    \toprule
    Method & \textbf{mIoU (\%)$\uparrow$} & \textbf{Precision$\uparrow$} & \textbf{Recall $\uparrow$} & \textbf{F1-score$\uparrow$} \\
    \midrule
   MAT~\cite{mat} & 47.5  & 0.82  & 0.54  & 0.56\\
   COS~\cite{cos} & 55.0  & 0.67 & 0.77 & 0.65  \\
   MotionGrouping~\cite{yang2021self} &16.2 &0.64 &0.19 &0.25\\
   AMD~\cite{liu2021emergence} &31.5 &0.42 &0.62 &0.45 \\
    ParticleSfM\cite{zhao2022particlesfm}* &60.4 & 0.78   & 0.74 &0.70   \\
    
   \midrule
    Ours (Baseline: Cotracker + motion segmentation) &50.7 &0.72 &0.66 &0.62 \\
    Ours (Dynamic-aware point tracking w/o depth) &49.8 & 0.59 & 0.83 & 0.61 \\
   Ours (Dynamic-aware point tracking w/ depth) & 54.9  & 0.73 & 0.71 &0.66 \\
   Ours (Dynamic-aware point tracking w/ consistent depth) & 53.9 & \textbf{0.80} & 0.62 & 0.65 \\
    \bottomrule
  \end{tabular}
  
  \label{tab:sintel_motion}
\end{table*}

\section{Method}
Our goal is to achieve smooth camera trajectories and consistent video depth with casual monocular videos. To achieve this, we present a dynamic-aware tracking any point method, termed DATAP, to estimate the point trajectories across videos with their visible and dynamic characteristics. Incorporating with DATAP, we formalize a concise, elegant, and robust pipeline of structure from motion in the wild. Fig.~\ref{fig:overview} outlines the pipeline of our method.

\subsection{Dynamic-Aware Tracking Any Point (DATAP)}
Given a casual monocular video $V = (I_t)_{t=1}^T$ containing $T$ RGB images $I_t \in \mathbb{R}^{3 \times H \times W}$ as input, DATAP aims to predict the trajectories $ P^j_t = (x^j_t, y^j_t) \in \mathbb{R}^{2}$, $ t =t^j,\dots,T $, $ j=1,\dots,N$, where $ t^j \in \{1, \dots, T\} $ of $N$ points across the video with the corresponding visibility $v^j_t \in \{0,1\}$ and dynamic labels $m^j_t \in \{0,1\}$. The visibility indicates whether the point is visible or occluded, and the dynamic label identifies whether the point is moving or stationary relative to the camera. Following CoTracker~\cite{karaev2023cotracker}, we assume that the state of the starting point of each track is visible $v^j_{t_j} = 1$ and dynamic $m^i_{t_j} = 1$. DATAP will obtain the estimates $(\hat P^j_t = (\hat x^j_t, \hat y^j_t), \hat v^j_t, \hat m^j_t)
$ of the location of the tracked points and their visibility and dynamics with the transformer network $ F: G \mapsto O $, where $G$ is the input tokens of the track, and $O$ is the output.

\textbf{Feature extraction and correlation.} For each RGB image $I_t \in \mathbb{R}^{3 \times H \times W}$ of the video $V = (I_t)_{t=1}^T$, we use a convolutional neural network to extract 4 layers of dense appearance features $\phi_s(I_t) \in \mathbb{R}^{d\times \frac{H}{s} \times \frac{W}{s}}$, $s=2,4,8,16$. The appearance features $Q_t^j \in \mathbb{R}^{d}$ of each track are initialized by sampling image features of the starting position over time and updated through the network. 

To compute the correlation between the track features $Q_t^j$ and the image features $\phi(I_t)$ surrounding the current estimate of track position $\hat P_t^j$, we adopt the dot products in RAFT~\cite{teed2020raft} to obtain the correlation volume. The correlation features $C_{t}^j$ are obtained through bilinear interpolation of the stacked inner products 
$ [C_t^j]_{s\delta} = \langle Q^j_t,~\phi_s(I_t)[\hat P_t^j / s + \delta]\rangle,$ where $s=2,4,8,16$ are the feature scales and $ \delta \in \mathbb{Z}^2$ are the offsets, which are within a 
radius of $r$ units: $\| \delta \| \leq r$.

\textbf{Input tokens.} Following CoTracker~\cite{karaev2023cotracker}, the input of the transformer is the tokens $G(\hat P, \hat v, \hat m, Q)$ which represent the position, visibility, dynamics, the appearance feature, and the correlation of the tracks. These tokens are concatenated with the positional embedding $\gamma$ of the start position $P^i_1$ and time $t$ of the track: $G^j_t = (\hat P^j_t - \hat P^j_1,~\hat v^j_t, \hat m^j_t,~Q^j_t,~C^j_t,~\gamma (\hat P^j_t - \hat P^j_1))+ \gamma(\hat P^j_1) + \gamma(t)$. 

\textbf{Iterative transformer updates.} The transformer update $F$ will be applied $K$ times to progressively update the estimates of the tracks from an initial token G. With each iteration, we can obtain the delta of the position $\Delta \hat P$ and the feature $\Delta Q$ of the tracks: $ O(\Delta \hat P, \Delta Q) = F(G(\hat P^{(m)}, \hat v^{(0)}, \hat m^{(0)}, Q^{(k)}))$, and the current estimate of position and feature are $\hat P^{(k+1)} = \hat P^{(k)} + \Delta \hat P $ and $ Q^{(k+1)} = Q^{(k)} + \Delta Q$, respectively. The transformer does not iteratively update the visibility mask $\hat v$ and dynamic mask $\hat m$ but performs the transformer update on the last iteration. The transformer is followed by an MLP and a sigmoid activation function $\sigma$ to update: $\hat v = \sigma(W Q) $ and $\hat m = \sigma(W Q) $, where $W$ is the learned weights of MLP. We find that such updates can better predict the visibility mask $\hat v$, but the dynamic mask $\hat m$ cannot be accurately obtained.

\begin{figure*}[htbp]
    \centering
    \includegraphics[width=\linewidth]{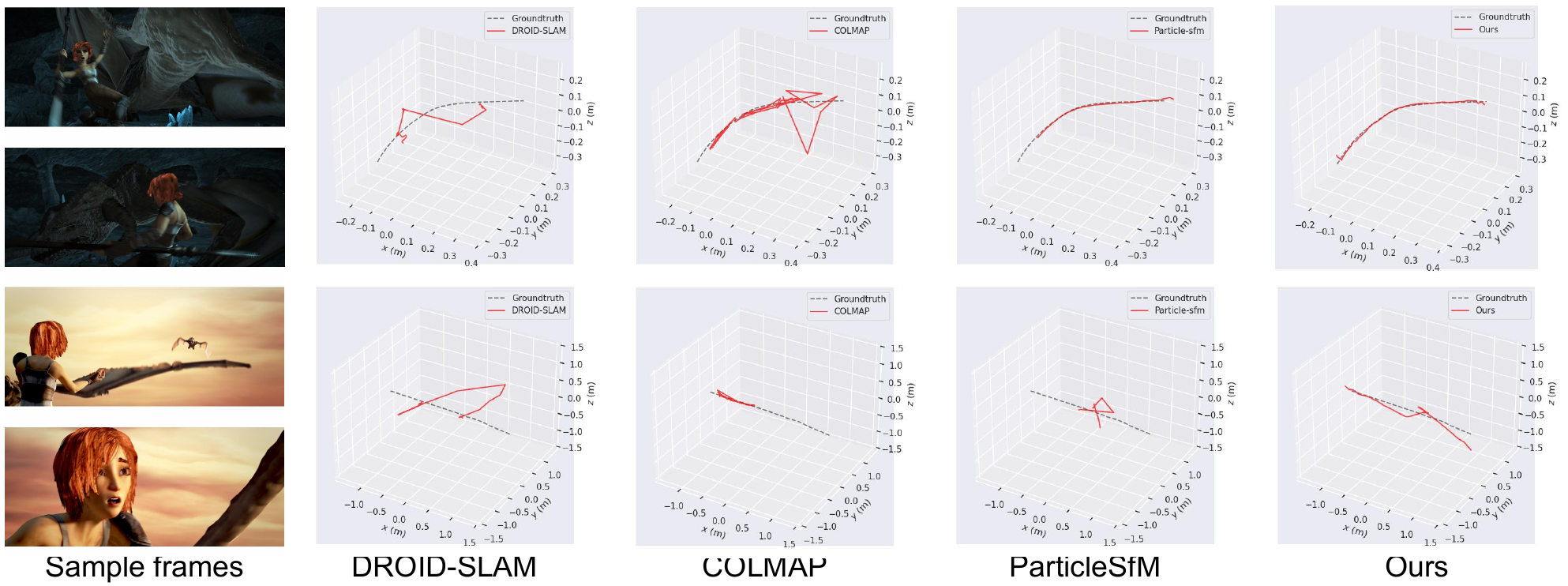}
    \caption{\textbf{Qualitative results of camera pose estimation on MPI Sintel dataset. Our method outperforms existing SOTA methods.} 
    }
    \label{fig:sintel}
\end{figure*}

\begin{table*}[t]
\centering
\caption{\textbf{Quantitative evaluations of camera pose on the MPI Sintel benchmark. Our method outperforms existing SOTA methods.} (Top: Successful subset of COLMAP, Bottom: Full set). 
  * means running with the source code. X means failure.
  }
\begin{tabular}{cccccc}
    \toprule
    & Method & \textbf{ATE (m)$\downarrow$} & \textbf{RPE Trans (m)$\downarrow$} & \textbf{RPE Rot (deg)$\downarrow$}\\
    \midrule
   & COLMAP~\cite{colmap} &0.145 &0.035 &0.550  \\
   & MAT~\cite{mat} + COLMAP &0.069 &0.024 &0.726  \\
COLMAP & Mask-RCNN~\cite{maskrcnn} + COLMAP  &0.109 &0.039 &0.605  \\
subset   & ParticleSfM\cite{zhao2022particlesfm}* &0.021 &0.007 &0.140  \\
   & Ours & \textbf{0.015} & \textbf{0.007} & \textbf{0.128} \\
  \midrule
     & COLMAP~\cite{colmap} &X &X &X \\
   
   & R-CVD~\cite{kopf2021rcvd} &0.360 &0.154 &3.443\\
 & Tartan-VO~\cite{tartanvo}  &0.290 &0.092 &1.303  \\
  Full  & DROID-SLAM~\cite{teed2021droid} &0.175 &0.084 &1.912  \\
 set  & ParticleSfM~\cite{zhao2022particlesfm}* &0.129 & \textbf{0.031} &0.535  \\
   & Ours & \textbf{0.104} & 0.037 & \textbf{0.306} \\
   
    \bottomrule
  \end{tabular}
  
  \label{tab:sintel}
\end{table*}

\textbf{Depth-aware trajectory motion segmentation.} 
Inspired by ParticleSfM~\cite{zhao2022particlesfm}, we introduce a depth-aware trajectory feature encoder and a decoder like OANet~\cite{oanet}} to disambiguate dynamic label prediction. Specifically, for each frame of the video, we use monocular depth estimation such as Midas~\cite{birkl2023midas} or consistent video depth such as NVDS~\cite{nvds} to obtain an initial depth estimate. Directly using 2D point tracking to predict dynamic labels will suffer from ambiguity. We normalize the relative depth of each frame to (0,1) and back-project it to 3D camera coordinates. For this reason, the trajectory of 2D point tracking can obtain sequential scene flow estimates. Referring to ParticleSfM~\cite{zhao2022particlesfm}, for the trajectory of the sliding window $L$, we concat the coordinates of the 2D trajectory, the coordinates of the 3D trajectory, the motion of the 2D trajectory, and the motion of the scene flow to form $L*10$ features. These features are first fed into 2 layers of MLP and then fed into a transformer module to obtain the encoded features. 

Following OANet~\cite{oanet}, the decoder first uses PointCN to obtain the local-global context features of the trajectory points, then uses softmax in the Diff Pool module to learn to cluster the input features, then performs spatial association on the clusters, and recovers each point through Diff Unpool contextual features. The features obtained by the Unpool layer are fed into several PointCN and followed by sigmoid activation, plus dynamic prediction of iterative transformer updates, to obtain the final dynamic mask. For more details, please refer to the OANet~\cite{oanet}.

\begin{table*}[t]
\centering
\caption{\textbf{Quantitative evaluations of camera pose on the TUM RGBD Dynamic benchmark. Our method outperforms existing SOTA methods.} (Top: Successful subset of ParticleSfM, Bottom: Full set). 
  }
\begin{tabular}{cccccl}
    \toprule
    & Method & \textbf{ATE (m)$\downarrow$} & \textbf{RPE Trans (m)$\downarrow$} & \textbf{RPE Rot (deg)$\downarrow$}\\ 
    \midrule
   
  ParticleSfM subset & ParticleSfM\cite{zhao2022particlesfm}* & 0.263	& \textbf{0.115}	& 11.554  \\
   & Ours & \textbf{0.193} (+26.62\%)	& 0.126	& \textbf{9.364}\\
  \midrule
   
  Full & ParticleSfM~\cite{zhao2022particlesfm}* & X & X & X\\
   subset & Ours &\textbf{0.185} &\textbf{0.122} & \textbf{10.209} \\
    \bottomrule
  \end{tabular}
  
  \label{tab:tum}
\end{table*}

\textbf{Supervision.} Dynamic-aware point tracking aims to estimate the trajectory, visibility, and dynamics of sampled points of the video. The loss function consists of three parts, namely the trajectory regression loss $L_{traj}$ within the sliding window $L$, the visibility cross-entropy loss $L_{vis}$, and the dynamic cross-entropy loss $L_{dyn}$. Among them, $L_{traj}$ calculates the $L1$ loss of the true trajectory and the predicted trajectory within the sliding window: 
$\mathcal{L}_{traj}(\hat{P}, P) = \sum_{j=1}^{J}\|\hat P^{(j)} -P^{(j)}\|$.
The visibility cross-entropy loss $L_{vis}$ is defined as: 
$\mathcal{L}_{vis}(\hat{v}, v) = \sum_{j=1}^{J} \operatorname{CE}(\hat v^{(j)}, v^{(j)})$.
Similarly, the dynamic cross-entropy loss $L_{dyn}$ is defined: 
$\mathcal{L}_{dyn}(\hat{m}, m) = \sum_{j=1}^{J} \operatorname{CE}(\hat m^{(j)}, m^{(j)})$.
The total loss is 
$L_{total} = \lambda_{traj} L_{traj} + \lambda_{vis} L_{vis} + \lambda_{dyn} L_{dyn}$. 

\begin{figure*}[htbp]
    \centering
    \includegraphics[width=\linewidth]{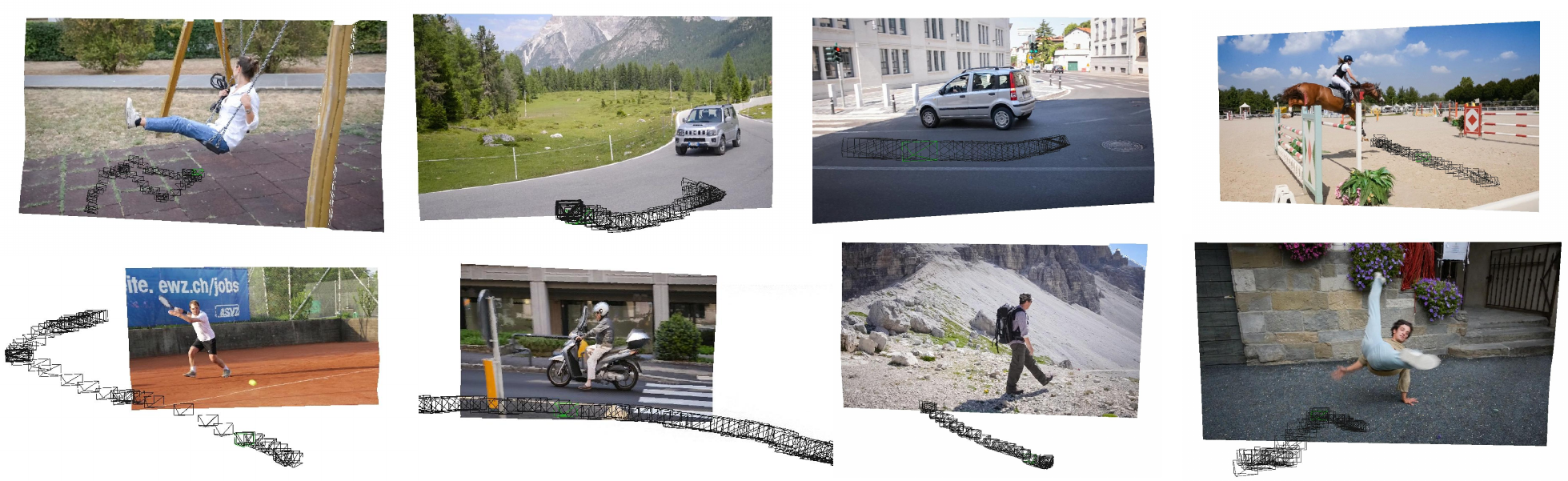}
    \caption{\textbf{Qualitative results of structure from motion on DAVIS dataset.} Our method can obtain smooth camera trajectories and entire point clouds of dynamic scenes.
    }
    \label{fig:davis}
\end{figure*}

\subsection{Structure from Motion with DATAP} 
Since DATAP can obtain dynamic-aware point-tracking video correspondence, we can directly perform nonlinear geometric optimization on point tracking to obtain camera poses and point clouds. Inspired by global SfM methods~\cite{sweeney2015theia}, we build a global SfM pipeline using dense point tracking. Specifically, trajectories marked as visible and static in point tracking are first extracted, and then translation averaging~\cite{ozyesil2015robust} and rotation averaging~\cite{chatterjee2013efficient} commonly used in global SfM pipelines are performed to obtain initial camera pose estimates. Then we apply global bundle adjustment over the selected point trajectories during the triangulation stage. Since there are scale differences in the point clouds obtained by SfM and depth estimation methods, we first project the consistent video depth into point clouds, then scale-align the point clouds marked as static labels with the point clouds of global SfM. The point cloud from depth estimation marked as dynamic is converted and fused to the SfM point cloud to obtain a complete point cloud of the entire dynamic scene.

\section{Experiment}
We conduct a comprehensive evaluation of DATAP-SfM in standard dynamic scenes such as MPI Sintel dataset~\cite{sintel}, and dynamic indoor scenes such as TUM RGBD dynamic sequence~\cite{tumrgbd} 
to evaluate its performance and generalization ability in complex challenge scenarios and compare it with many state-of-the-art SLAM or SfM methods. We also conduct qualitative verification on casual videos in the wild such as DAVIS~\cite{davis}.

\subsection{Implementation Details}
Following ParticleSfM, we mainly train our DATAP model on the Flyingthing3D dataset~\cite{brox2010large}. We first train the trajectory prediction task with lr=5e-4 for 50 epochs, then fix the trajectory prediction to train the depth-aware dynamic prediction for 30 epochs, and then finetune the whole model for 30 epochs until the network converges. The model was trained on 4 RTX3090s for 2 days.

\subsection{Dataset}
\textbf{MPI Sintel dataset.} 

\textbf{TUM RGBD dynamic sequences.} The TUM RGBD dataset~\cite{tumrgbd} is a benchmark for evaluating SLAM or SfM algorithms on different challenges. To evaluate the performance of our method in dynamic scenes, we selected 9 sequences containing dynamic objects, such as moving people.

\paragraph{ScanNet.} The ScanNet dataset~\cite{scannet} is a benchmark for evaluating SLAM or SfM algorithms in indoor static scenes. It involves many challenges, such as weak textures, illumination changes, pure camera rotation, etc. To further verify the robustness of our method in static scenes, we follow ParticleSfM and select the first 20 sequences of the test set for evaluation.

\textbf{DAVIS.} The DAVIS dataset~\cite{davis} is a benchmark for evaluating video object segmentation and tracking without ground-truth camera poses. It contains many challenges, such as multi-object occlusion, complex motion, motion blur, etc. To further demonstrate the generalizability of our method, we qualitatively visualize the effect of our method.

\subsection{Evaluation metrics} 

Since our method is performing structure from motion of dynamic scenes, we mainly evaluate the results of motion segmentation and pose estimation.

\textbf{Motion Segmentation.} Motion segmentation aims to analyze whether the pixels of a video sequence are moving relative to the camera. We first calculate the ground-truth dynamic labels based on the forward and backward ground-truth optical flow of the dataset. Then calculate the difference between our trajectory prediction dynamic label and the ground-truth value. We follow commonly used evaluation metrics for motion segmentation, such as prediction, recall, F1 score, and IoU. It is worth noting that in the SfM method of dynamic scenes, the precision metric of motion segmentation is more important than other metrics. 

\textbf{Pose Estimation.} We follow the standard pose evaluation metrics of visual odometry or SLAM: relative pose error (RPE) and absolute trajectory error (ATE), where RPE measures the relative pose error of the pair frames, including relative rotation error (RPE Rot) and relative translation error (RPE Trans). ATE measures the root mean square error between the predicted camera pose and the ground truth. Since the scene scale is unknown, during the evaluation process, we adopt evo~\cite{grupp2017evo} to align the predicted pose with the ground truth pose for fair comparisons.

\subsection{Comparison on MPI Sintel dataset} 
The MPI Sintel dataset~\cite{ButlerSintel2012} is a synthetic natural video sequence that contains 23 complex challenging scenes such as highly dynamic, motion blurred, non-rigid motion, etc. Following ParticleSfM~\cite{zhao2022particlesfm}, we removed sequences that were ineffective for evaluating monocular camera poses, such as static camera motion, leaving 14 sequences for comparison. 
We compare our method with the classic feature-point SfM method COLMAP~\cite{colmap} and its variants and state-of-the-art deep learning methods~\cite{teed2021droid, rcvd, tartanvo}. The quantitative pose estimation results in Table~\ref{tab:sintel} show that COLMAP and its variants can only perform pose estimation on certain sequences. The learning-based state-of-the-art method, such as DROID-SLAM~\cite{teed2021droid} performs poorly in dynamic scenes and struggles to obtain accurate camera trajectories. While the recent ParticleSfM performs well on most scenes, our SfM method based on dynamic-aware point tracking far outperforms them, with a 19.37\% improvement in ATE on all datasets, and a 28.57\% improvement in ATE on the COLMAP subset. 
Fig.~\ref{fig:sintel} shows the qualitative pose estimation comparison results on the Sintel dataset. We can observe that even in some challenging scenarios, our method can obtain more accurate pose estimation results than the existing Sota method.




\subsection{Comparison on TUM RGBD Dynamic dataset}
The TUM RGBD dataset~\cite{tumrgbd} is a benchmark for evaluating SLAM or SfM algorithms on different challenges. To evaluate the performance of our method in dynamic indoor scenes, we selected 9 sequences containing dynamic objects, such as moving people. Since ParticleSfM often performs better on dynamic scenes, we choose ParticleSfM as the main comparison. By running the open source code of ParticleSfM, experiments show that ParticleSfM will have a system failure in the 9 datasets of TUM, and our method can solve the camera pose, which shows the robustness of our method. In the subset of ParticleSfM, our method has a 26.62\% improvement on ATE.

\subsection{Qualitative evaluation on in-the-wild videos}
The DAVIS dataset is a benchmark for evaluating video object segmentation and tracking without ground-truth camera poses. It contains many challenges, such as multi-object occlusion, complex motion, motion blur, etc. To further demonstrate the generalizability of our method, we select 15 sequences from the DAVIS dataset and qualitatively visualize the effect of our method. Experiments show that COLMAP can only run 10 of DAVIS's 15 sequences, while ParticleSfM can only run 8 sequences. They struggle to obtain satisfactory pose estimation. Fig.~\ref{fig:teaser} demonstrates that compared with COLMAP and ParticleSfM, our method can obtain more accurate pose estimation and complete point cloud of the entire dynamic scenes. We also show more qualitative results in Fig.~\ref{fig:davis}.

\begin{table*}[t]
\centering
\caption{\textbf{Ablation study of camera pose on the MPI Sintel benchmark.} (Left: Full set, Right: Successful subset of COLMAP
  ). 
  }
\begin{tabular}{lccc}
    \toprule
    Method & \textbf{ATE (m)$\downarrow$} & \textbf{RPE Trans (m)$\downarrow$} & \textbf{RPE Rot (deg)$\downarrow$}\\
    \midrule
   COLMAP~\cite{colmap} & X / 0.145 & X / 0.035 & X / 0.550  \\
    ParticleSfM\cite{zhao2022particlesfm}* & 0.132 / 0.021 & 0.038 / 0.006 & 0.915 / 0.140  \\
   \midrule
    Ours (Baseline: CoTracker + motion segmentation + SfM) & 0.128 / 0.023 & \textbf{0.032} / 0.009   & \textbf{0.228} / 0.107\\
    Ours (Dynamic-aware point tracking w/o depth + SfM) & 0.138 / 0.023	& 0.045 / 0.015	& 0.472 / 0.197  \\
   Ours (Dynamic-aware point tracking w/ depth + SfM) & \textbf{0.104} / 0.023	&0.037 / 0.013	& 0.306 / 0.143 \\
   Ours (Dynamic-aware point tracking w/ consistent depth + SfM) & 0.117 / \textbf{0.015} & 0.039 / \textbf{0.007} & 0.376 / \textbf{0.128} \\

    \bottomrule
  \end{tabular}
  
  \label{tab:sintel_pose}
\end{table*}

\subsection{Ablation Study}
We conduct ablation experiments on the network design of our method in motion segmentation and pose estimation. ParticleSfM constructs point trajectories by linking paired optical flows, which undoubtedly brings cumulative errors. 

\textbf{Baseline.} Our baseline directly uses the pre-trained point tracking method to replace ParticleSfM's point trajectory construction, plus ParticleSfM's point trajectory motion segmentation and global bundle adjustment. It can only obtain results comparable to ParticleSfM in pose estimation, as shown in Tab.~\ref{tab:sintel_pose}, but cannot obtain better results. Our analysis: the accuracy of motion segmentation is just comparable to that of ParticleSfM, shown in Tab.~\ref{tab:sintel_motion}, and the pre-trained point tracking method~\cite{karaev2023cotracker} cannot be well adapted to the motion segmentation of ParticleSfM. 

\textbf{Dynamic-aware point tracking w/o depth.} We construct an e2e dynamic-aware point tracking without depth estimation. Experiments show that end-to-end methods without depth prior cannot effectively distinguish motion segmentation and suffer from ambiguity.

\textbf{Dynamic-aware point tracking w/ depth.} We observe from Tab.~\ref{tab:sintel_motion} and Tab.~\ref{tab:sintel_pose} that 
 with the depth prior, the accuracy of motion segmentation and camera pose has been greatly improved. 

\textbf{Dynamic-aware point tracking w/ video depth.} Compared with monocular depth prior, the precision of motion segmentation can be further improved. As shown in Fig.~\ref{fig:sintel_motion}, our method obtains more accurate motion segmentation, while ParticleSfM may misidentify static regions as dynamic.

\section{Conclusion}
We have presented a structure-from-motion method with dynamic-aware point-tracking for accurate pose estimation. Our method can obtain smooth camera trajectories and entire point clouds of dynamic scenes for casual videos in the wild and outperforms existing SfM and SLAM methods in dynamic scenes.

\textbf{Limitation.} Although our method can perform robust pose estimation and consistent video depth estimation in dynamic scenes, it cannot operate like a real-time SLAM system, even if we adopt a sliding window-based point tracking mechanism. Developing an efficient dynamic-aware point-tracking method is a promising direction. Exploring large-scale and diverse Internet videos to train point-tracking methods will further improve the robustness. We leave it as future work.

\section{Broader Impact.} This paper proposes a method to estimate smooth camera trajectories and obtain the entire dense point cloud of dynamic scenes from arbitrary monocular videos. This method can provide initial accurate pose and depth for dynamic scene NeRF reconstruction and editing such as film and television production, etc.


\bibliography{DATAP-SfM}
\bibliographystyle{icml2024}

\end{document}